# Bayesian Poker

Kevin B. Korb, Ann E. Nicholson and Nathalie Jitnah
School of Computer Science and Software Engineering
Monash University, Clayton, VIC 3168 AUSTRALIA
{korb,annn,njitnah}@csse.monash.edu.au

## Abstract

Poker is ideal for testing automated reasoning under uncertainty. It introduces uncertainty both by physical randomization and by incomplete information about opponents' hands. Another source of uncertainty is the limited information available to construct psychological models of opponents, their tendencies to bluff, play conservatively, reveal weakness, etc. and the relation between their hand strengths and betting behaviour. All of these uncertainties must be assessed accurately and combined effectively for any reasonable level of skill in the game to be achieved, since good decision making is highly sensitive to those tasks. We describe our Bayesian Poker Program (BPP), which uses a Bayesian network to model the program's poker hand, the opponent's hand and the opponent's playing behaviour conditioned upon the hand, and betting curves which govern play given a probability of winning. The history of play with opponents is used to improve BPP's understanding of their behaviour. We compare BPP experimentally with: a simple rule-based system; a program which depends exclusively on hand probabilities (i.e., without opponent modeling); and with human players. BPP has shown itself to be an effective player against all these opponents, barring the better humans. We also sketch out some likely ways of improving play.

## 1 INTRODUCTION

Poker is an ideal vehicle for testing automated reasoning under uncertainty. Poker is a game intermediate between chess and chance: there are important opportunities for the exercise of strategic planning, tactical skills and the astute observation and learning of the abilities and tendencies of opponents; and there are important elements of uncertainty, making all of learning, planning and execution difficult. Uncertainty enters through the direct introduction of physical randomness by shuffling; it also enters into the game through the player's incomplete knowledge of the state of opponents' hands. In addition, poker is famously a game of psychology — effective modeling of one's opponents and effective bluffing are critical for success. Finally, poker is a game all about betting, and betting behaviour is the foundation and origin of our most powerful formalization for coping with uncertainty, namely probability theory, as well as our most powerful philosophical theory of uncertainty, namely Bayesianism (Ramsey, 1931). Considering all of these factors, it is somewhat surprising, ten years after Judea Pearl's seminal presentation of effective computational models of probabilistic reasoning (Pearl, 1988) that poker is only now beginning to be studied as an application domain. Here we present one such implementation of five-card stud poker using Bayesian networks, begun with (Jitnah, 1993) and brought to a competitive level of play recently.

There has been some prior work on automated poker play. Findler (1977) used a combination of a probabilistic assessment of hand strength with the collection of frequency data for opponent behaviour to support the refinement of the models of opponent. The frequency data were collected separately for each opponent and each round of play (in his "Bayes 4" model).

Koller and Pfeffer (1997) present a framework, based on an augmented game tree, for generating and solving imperfect information games such as poker. Although their algorithm finds optimal randomized strategies in time polynomial with respect to the game tree, the size of the game tree for full poker is still prohibitive. They suggest abstracting the problem by partitioning the game tree into clusters. Our representation is a Bayesian network model, rather than a game tree, however we reduce the problem size using abstraction by using hand types.

Most recently, Billings *et al.* (1998) have investigated the automation of the poker game Texas Hold'em. They combine probability tables for current hand strength, the probability of a hand improving during play, frequency data for opponent behaviour, and playing heuristics. Their work models opponents in a weighting scheme relating hand strength to betting behaviour. Our work similarly uses frequency data to learn by updating the conditional probability of betting behaviour given the state of the opponent's hand, and it also models opponents, but does so more explicitly as a subnet in a Bayesian network. Hold'em is an interesting poker variation, but has the drawback that all known (up) cards are shared by all the players. This means that the potential of any hand is (in some sense) assessed against an average hand, since nothing is known explicitly about opponents' hands, except what has been revealed by the opponents' behaviour. Stud poker, which we have chosen to study, reveals partial information about each hand during play, requiring that information to be combined with observation of behaviour to assess the potential of an opponent's hand.

In Section 2, we describe five-card stud poker and how a probability of winning is optimally used to select an action given the current size of the pot. In Section 3 we describe a simple Bayesian network which compactly represents two-person poker and provides an estimate of the probability of winning after each card is dealt. In Section 4 we discuss the betting (playing) curves, functions for probabilistically converting pot odds and a winning probability into an action, and our procedure for optimizing them. We also discuss bluffing as a method for further camouflaging hand strength strategically. Then in Section 5 we describe the experiments we have conducted to test the effectiveness of our program. Finally, we discuss some likely ways of improving the program (Section 6).

## 2   FIVE-CARD STUD POKER

### 2.1   THE GAME

In five-card stud poker, after an ante (an initial fixed-size bet), players are dealt a sequence of five cards, the first down (hidden) and the remainder up (available for scrutiny by other players). Players bet after each up card is dealt, in a clockwise fashion, beginning with the best hand showing. The first player(s) may PASS — make no bet, waiting for someone else to open betting. Bets may be CALLED (matched) or RAISED, with up to three raises per round. Alternatively, a player facing a bet may FOLD her or his hand, i.e., drop out for this hand. After the final betting round, among the remaining players the one with the strongest hand wins in a "showdown". The strength of poker hand types is strictly determined by the probability of the hand type appearing in a random selection of five cards (see Table 1). Two hands of the same type are ranked according to the value of the cards (without regard for suits); for example, a pair of Aces beats a pair of Kings.

Table 1: Poker Hands: weakest to strongest

| Hand Type | Example | Probability |
|---|---|---|
| Busted | A♣ K♣ J◇ 10◇ 4♡ | 0.5015629 |
| Pair | 2♡ 2◇ J♠ 8♣ 4♡ | 0.4225703 |
| Two Pair | 5♡ 5♣ Q♠ Q♣ K♣ | 0.0475431 |
| Triple | 7♣ 7♡ 7♠ 3♡ 4◇ | 0.0211037 |
| Straight (sequence) | 3♣ 4♣ 5♡ 6◇ 7♠ | 0.0035492 |
| Flush (same suit) | A♣ K♣ 7♣ 4♣ 2♣ | 0.0019693 |
| Full House | 7♠ 7◇ 7♣ 10◇ 10♣ | 0.0014405 |
| Four of a Kind | 3♡ 3♠ 3◇ 3♣ J♠ | 0.0002476 |
| Straight Flush | 3♠ 4♠ 5♠ 6♠ 7♠ | 0.0000134 |

### 2.2   POT ODDS & PROBABILITIES

A basic decision facing any poker player, given that a bet is on the table, is whether to call or fold one's hand (ignoring the possibility of raising *pro tem*). Assuming probability $p$ of winning the pot if the hand is played to a showdown, $n-1$ opponents remain in the game and an expected cost $k$ of reaching the showdown, then the threshold probability of winning required to make the decision to call a bet can be worked out from the *pot odds*. Letting $f$ be the size of the final pot and $c$ the current size of the pot, then, following Zadeh (1974):[1]

$$\text{pot odds}_Z \;=\; \frac{k}{f-k} \qquad (1)$$

$$\simeq \;\frac{k}{c+(n-1)k} \qquad (2)$$

This assumes that the final size of the pot minus the current player's future contribution will be the current size of the pot plus equal contributions by all other players. The calling threshold $\theta_Z$ identifies the probability of winning at which the expected values of calling a bet versus folding are equal. By the standard relation between probability and odds, $\theta_Z$ is then:

$$\theta_Z \;=\; \frac{\text{pot odds}}{1+\text{pot odds}} \qquad (3)$$

---
[1] One approximation here is that this ignores any players who contribute to the pot in the future but fail to stay to the showdown. In two-player games, that case does not arise. Billings *et al.* (1998), incidentally, give (1) incorrectly as pot odds $= k/f$ (in our notation; see p. 496).

$$= \frac{k}{c+nk} \quad (4)$$

Zadeh's formula (2) for pot odds is not quite right, however, because the cost $k$ of reaching the showdown is not the same for all players. Since the contemplated scenario is matching someone else's bet, it will clearly cost the decision maker one more bet to reach the showdown than at least one opponent. As a better approximation, we can assume that the current player is in the middle of the table — i.e., midway between the first bettor and the last person called upon to match that bet. In that case, (2) should be replaced by:

$$\text{pot odds} = \frac{k}{c+(n-1)k-\left(\frac{n-1}{2}\right)u} \quad (5)$$

where $u$ is the size of the betting unit. Given $u = 1$ and the restriction to a single opponent ($n = 2$), which currently applies to BPP, the denominator becomes $c + k - 1/2$. However, this formula reflects a false uncertainty as to BPP's position at the table: since the basic decision we are considering is whether to call, and not to bet, the correct formula is:

$$\text{pot odds} = \frac{k}{c+k-1} \quad (6)$$

So, finally, by (3)

$$\theta = \frac{k}{c+2k-1} \quad (7)$$

If we can compute an accurate probability of winning, then by comparison with $\theta$ we can make reasonable decisions about whether to call or fold. By extension — considering the degree to which our estimated winning probability exceeds that threshold when it does — we can make reasonable decisions about bets and raises as well. (Adjustments for randomization and bluffing are discussed below.) The main problem is to assess winning chances accurately. This clearly is not *simply* a matter of the probability that the hand you have now, if dealt out to the full five cards, will end up stronger than your opponent's hand as far as you can see it (i.e., its upcards), if it is also dealt out. For testing purposes, however, we do use just such a program as an opponent to BPP: at each decision point it simulates enough hand completions to have an excellent estimate of the probability of winning, going by the current known cards alone, and it uses this probability alone for decision making.[2]

Such a pure combinatorial probability is clearly of interest, but it ignores a great deal of information that good poker players rely upon. It ignores the "tells" that some poker players have: facial tics, pupil dilation, fidgeting, etc., revealing abnormally strong or weak hands; it also ignores current opponent betting behaviour and the past association between betting behaviour and hand strength. Without a robot's sensory apparatus, our program is in no position to deal with tells, but it does account for current betting behaviour and learns from the past relationship between opponents' behaviour throughout the game and their hand strength at showdown. The learning can be particular to a specific opponent or it can combine information across opponents.

## 3  A BAYESIAN NETWORK FOR POKER

### 3.1  NETWORK STRUCTURE

BPP uses a simple Bayesian network structure (Figure 1) for modeling the relationships between current hand type, final hand type and the behaviour of the opponent. We maintain such a network structure for each of the four rounds of play (the betting rounds after two, three, four and five cards have been dealt). The number of cards involved in the current and observed hand types, and the conditional probability matrices for them, vary for each round: in effect, then, we use four distinct Bayesian networks to govern play.

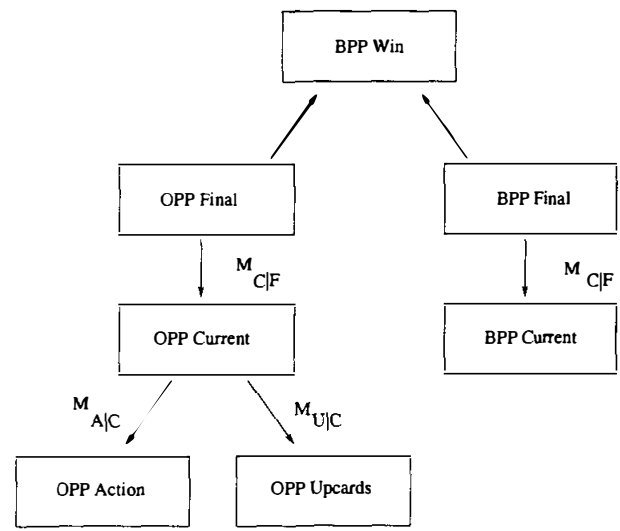

Figure 1: Bayesian Poker network

The node *OPP Final* represents the opponent's final hand type, while *BPP Final* represents BPP's final hand type; that is, these represent the hand types they will have after all five cards are dealt. Whether or not BPP will win is the value of the Boolean variable *BPP Win*; this will depend on the final hand types, deterministically when the hand types are different, and probabilistically when the hand types are equal. BPP's final hand is an observed variable after the final card is dealt, whereas its opponent's final hand type is observed only after play ends in a showdown. At any given stage in the game, BPP's current hand type is represented by the node *BPP Current* (an observed variable), while *OPP Current* represents its opponent's current hand type. Since BPP cannot observe its opponent's current hand type, this must be inferred from the information available: the opponent's upcard hand type, represented by node *OPP Upcards*, and the opponent's actions, represented by node *OPP Action*. Note that until the final round *BPP Current*, *OPP Current* and *OPP Upcards* represent partial hand types (e.g., three cards to a flush).

---

[2] This program is due to Jon Oliver.

In order to keep the network structure a simple polytree, we have made certain assumptions. In particular, the current network structure assumes that the final hand types are independent and that the opponent's action depends only on its current hand; we discuss the relaxation of these assumptions in Section 6.

## 3.2 HAND TYPES

The nodes representing hand types were initially given values which divided hands into the nine categories of final hand reported in Table 1 (Jitnah, 1993). This produced a level of play comparable to a weak amateur. Since any busted hand, for example, is treated as equal to any other, BPP would bet inappropriately strongly on middling busted hands and inappropriately weakly on Ace-high busted hands. The lack of refinement of paired hands also hurt its performance. In principle we could provide a different hand type to each poker hand that has a distinct value, since there are only finitely many of them. That finite number, however, is fairly large from the point of view of Bayesian net propagation; for example, there are already 156 differently valued Full Houses. We opted for a modest additional refinement, moving from 9 hand types to 17 types, subdividing the busted hands into busted-low (9 high or lower), busted-medium (10 or J high), busted-queen, busted-king and busted-ace; we subdivided pairs likewise. This appears to be sufficient to achieve quite good results in five-card stud amateur play. If we were to move to games involving more dealt cards (e.g., seven-card stud) or games with wild cards, we would need to further refine the higher categories of hand, as they would appear more often.

## 3.3 CONDITIONAL PROBABILITY MATRICES

There are four action matrices $M_{A|C}$ corresponding to the four rounds of betting. These report the conditional probabilities per round of passing or calling versus betting or raising, given the opponent's current hand type. BPP adjusts these matrices over time, using the relative frequencies of these behaviours. Since the rules of poker do not allow the observation of hidden cards unless the hand is held to showdown, these counts are made only for such hands. This is likely to introduce some selection bias into the estimates of conditional probabilities, but we have not yet attempted to determine the nature of the bias.

The four matrices $M_{U|C}$ give the conditional probabilities of having a given hand showing on the table when the opponent's hand (including the hidden card) is of a certain type. The four matrices $M_{C|F}$ (used for both *OPP Current* and *BPP Current*) give the conditional probability for each type of partial hand given that the final hand will be of a particular kind. These matrices were estimated by dealing out 10,000,000 hands of poker.

## 3.4 BELIEF UPDATING

Belief updating is done by standard Bayesian net propagation rules (Pearl, 1988). Given evidence for *BPP Current*, *OPP Upcards* and *OPP Action*, belief updating produces belief vectors for both players final hand types and, most importantly, a posterior probability of BPP winning the game.

## 4 RANDOMIZATION

### 4.1 BETTING CURVES

As described above, the calling threshold $\theta$ identifies the probability of winning at which the expected values of calling a bet versus folding are equal. Probabilities substantially higher than $\theta$ should in general lead to bets and raises; probabilities much lower to passes and folds. However, if a player invariably bets strongly given a strong hand and weakly given a weak hand, other players will quickly learn of this association; this will allow them to better assess their chances of winning and so to maximize their profit at the expense of the more predictable player. Therefore, we use betting curves, such as that of Figures 2, to randomize the actions of BPP in a way dependent upon the probability of winning. The horizontal axis shows the difference between that probability and $\theta$; the vertical axis is the (unnormalized) probability of a given action: fold, call or bet (raise). The normalized probabilities are used to stochastically select an action in any situation.

The playing curves are generated by the following exponential functions (where $d$ is the winning probability minus $\theta$ and $f$ is a parameter adjustable for each round of play):

$$\text{bet/raise prob} = \frac{1}{1 + e^{-8(d-f_b)}} \quad (8)$$

$$\text{fold prob} = \frac{1}{1 + e^{8(d+f_f)}} \quad (9)$$

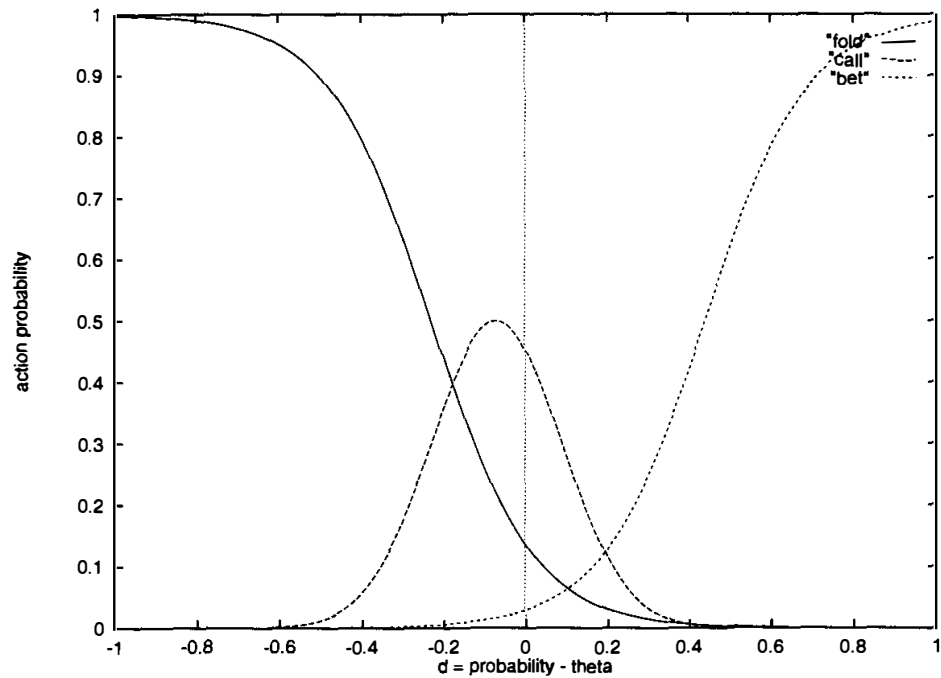

Figure 2: Final round betting curves (fold, call, bet). The horizontal axis is the difference between the probability of winning and $\theta$; the vertical axis is the (unnormalized) probability of the action.

$$\text{call prob} = \frac{e^{-20(d+f_c)^2}}{2} \quad (10)$$

Ideal $f$ parameters will select the optimal horizontal displacement of each curve relative to $d$ and thereby the optimal balance between conservative and aggressive play. For example, if the folding curve is shifted to the right relative to the calling curve, more conservative play will result, with even moderately strong hands perhaps being dropped. Or if the betting/raising curve is shifted to the left, more aggressive play will result.

In order to find good $f$ parameters, we employed a stochastic search of the parameter space when running BPP against the rule-based opponent. Since the space being searched is 12 dimensional (three types of curves, four each for the rounds of play) and the score function is highly noisy (wins/losses in actual poker play), it is not clear that the search for optimality was successful. Nevertheless, the curves produced by our stochastic search appear to provide a reasonable answer to such questions as how much greater the probability of winning must be over the threshold for active bets and raises to be rewarded. Their use also provides good camouflage for playing behaviour by their introduction of random play.

One apparent anomaly is that the point at which the probability of folding equals the probability of calling should occur when the probability of winning is equal to $\theta$ (i.e., at 0 on the horizontal axis of Figure 2).

We believe the explanation is that our estimate of the pot odds (being dependent upon an estimate of the expected cost to a showdown) is inexact. If so, then when we optimize the playing curves, the optimization process will compensate for the estimation error by displacing the playing curves. Since the calling curve is displaced to the left, this suggests that $\theta$ is being overestimated.

### 4.2 BLUFFING

Bluffing is the intentional misrepresentation of the strength of one's hand. You may over-represent the strength of your hand (what is commonly thought of as bluffing), in order to chase opponents with stronger hands out of the round. You may equally well under-represent the strength of your hand ("sandbagging"), in order to retain players with weaker hands and relieve them of spare cash. These are tactical purposes behind almost all (human) instances of bluffing.

On the other hand, there is an important strategic purpose to bluffing, as von Neumann and Morgenstern (1953) pointed out, namely "to create uncertainty in [the] opponent's mind" (pp. 188-9). In BPP this purpose is already partially fulfilled by the randomization introduced with the betting curves. However, that randomization occurs primarily at the margins of decision making, when one is maximally uncertain whether, say, calling or raising is optimal over the long run of similar situations. Bluffing is not restricted to situations where the optimal normal action is uncertain; the need is to disguise from the opponent what the situation is, whether or not the optimal response is known to the player. Hence, bluffing is desirable for BPP as an action in addition to the use of randomizing playing curves. Since BPP's randomized play already results in some apparent bluffing, the probability with which BPP ought to bluff outright should be somewhat lower than otherwise. BPP currently bluffs (by over-representation) in the last round of betting with a low probability (5%).

## 5   EXPERIMENTATION

### 5.1   OPPONENTS

**A Simple Probabilistic Opponent**

The probabilistic player, as mentioned in § 2.2, estimates its winning probability for its current hand by taking a large sample of possible continuations of its own hand and its opponent's hand. Since our experimental intention was to compare the effectiveness of using this combinatorial probability versus the Bayesian net probability that incorporates the opponent's behaviour, once the combinatorial player computed its probability we used the same algorithm to complete its play as does BPP — i.e., the probabilistic opponent used the pot odds in the same way as BPP to determine its betting behaviour.

**A Rule-Based Opponent**

The rule-based system which we used as another opponent for BPP is described in Figure 4. These rules incorporate plausible maxims for play, for example, that you should generally fold your hand if it is already beaten by what's showing of your opponent's hand (i.e., its upcards).

**Human Opponents**

People who had experience playing poker were invited to try themselves against BPP.

### 5.2   RESULTS

The cumulative performance of BPP against the three different varieties of opponent we have tested against is shown in Figure 3. BPP is clearly outperforming the two automated opponents ("prob" and "rules"), with the discrepancy in their results suggesting that the rule-based system is superior overall to the probabilistic player (although we have not yet run them against each other to verify that). The "humans" recorded in Figure 3 shows the combined record of various people

```
If BPP-up > OPP-up
    If OPP > BPP-up
        If OPP-hand-type > BPP-up-hand-type
            Then BET/RAISE (90%), CALL/PASS (10%)
            Else BET/RAISE (80%), CALL/PASS (20%)
    Else {BPP-up > OPP}
        FOLD/PASS (85%), CALL/PASS (15%)
Else {OPP-up ≥ BPP-up}
    If BPP is betting
        Then If OPP-hand-type > BPP-up-hand-type
            Then RAISE (85%), CALL (15%)
            Else CALL (85%), FOLD (15%)
    Else BET.
```

Figure 4: Rule-based opponent. OPP-up is the rule-based system's hand showing on the table (likewise BPP-up); OPP is the opponent's (full) hand.

who took up an invitation to play BPP via telnet.[3] The evident variation in the human record is likely due to their different playing abilities. Given more play against the different individuals we would expect consistent trends to emerge. The "experienced" player is someone who has frequently and successfully played amateur poker; this player ended up with a net gain (loss to BPP) of only 63 betting units after about 450 games.

**Significance Results**

In order to confirm our impressions of relative performance from Figure 3 we conducted two-tailed significance tests of the null hypothesis that BPP's true expected winnings are zero against each opponent. BPP's performance was significantly different from zero against both the probabilistic opponent ($t = 15.8; p \leq 0.01$) and the rule-based opponent ($t = 7.84; p \leq 0.01$). Its performance against both sets of humans was not significantly different from zero.

**Learning**

During these tests the conditional probability matrices $M_{A|C}$ were learned individually per opponent, starting from a default assumption that aggressive play (betting, raising) was as likely as conservative play (passing, calling). The results of Figure 3 do not appear to show any large learning effect: other than the human performances, which do not reflect a single level of play, the slopes of BPP's cumulative winning record do not appear to change substantially after the first 200 games. However, since each showdown contributed equally to the updates of the matrices, the most learning should occur in the early stages of play with an opponent. In order to check whether learning was occurring, we performed a difference of means significance test for each opponent comparing early performance (the first 200 games) with late performance (the last 200 games). BPP's performance against the probabilistic opponent was significantly different ($t = 3.25; p \leq 0.01$), but showed no significant improvement against any of the other players.

---

[3]The automated opponents were run considerably longer than shown in Figure 3, but without any interestingly different results.

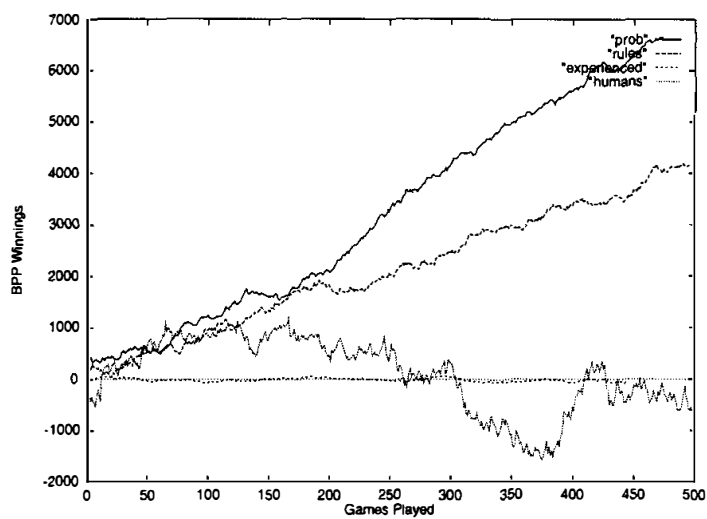

Figure 3: Bayesian Poker Program's cumulative performance

## 6 FUTURE WORK

### 6.1 IMPROVEMENTS

During the work on this project we have noted a variety of methods of improving the Bayesian network and/or its use in poker. We list some of the more promising ideas here.

**Refine action nodes.** Currently BPP merges the actions CALL and PASS, and again BET and RAISE. This is possible because the paired playing situations are mutually exclusive. However, it is unlikely that given the same winning probability and pot odds each dual is as good a choice as the other. Therefore, they should be separated and given independent playing curves.

**Further refinement of hand types.** It's clear that the hand types need further refinement to reach anything like expert play. It is not clear how far to go.

**Improve network structure.** A link should be added from *BPP Final* to *OPP Final* (or vice versa). This will allow knowledge of the cards of one hand (observed or inferred) to be employed in determining the unknown cards of the other. The program should also take advantage of the difference between BPP's upcards and its full hand. This could involve adding a node for BPP parallel to *OPP Upcards*. The point is that when one's strong cards are showing on the table, there is no reason to bet coyly; on the contrary, it is advantageous to make opponents pay for the opportunity of seeing any future cards by betting aggressively. On the other hand, when one's strongest card is hidden, aggressive betting can drive opponents out of the game prematurely. However, it's not clear how to develop the Bayesian network to allow for this. A simpler and more direct approach would be to divide the playing curves, which govern aggressive vs. conservative betting behaviour, so that one set applies when the hand type is already fixed by one's upcards (i.e., the strong cards are showing) and the other when the hand type is hidden. By optimizing these separately, the optimal behaviour described above should be discovered.

**Add bluffing to the opponent model.** Just as for BPP, the conservativeness or aggressiveness of an opponent's play, which is learned and captured by recalibrating the matrices relating *OPP Current* and *OPP Action*, does not fully describe the bluffing behaviour of the opponent. A plausible extension would be to add an opponent bluffing node which is a parent of *OPP Action* and a child of *OPP Current* and *BPP Current* (since the latter gives rise to BPP's upcards and behaviour, even though they are not explicitly represented).

**Improved Learning of Opponent Model.** We did not get much of a learning effect. We believe this is largely due to weaknesses in the representational power of the opponent model, limiting the consequences of learning. By refining the action nodes and adding modeling of bluffing, we would expect learn-ing applied to the conditional probability matrices relating the nodes of the opponent model to be more effective. Another structural change important for improved learning and play would be to make *OPP Action* a child node of a new *BPP Upcards* node, so that what the opponent observes of BPP's hand would jointly condition his or her behaviour.

## 6.2 MORE CHALLENGING POKER

In the future we would like to make the poker-playing environment more challenging, in particular introducing multi-opponent games and allowing table stakes games. In multiple opponent games it will be more important to incorporate the interrelations between what is known of different player's hands and the node representing their final hands. In table stakes any player may bet or raise as much money as is jointly available at the time to that player and any remaining opponent. This makes the precise computation of winning probabilities rather more critical. Especially, it will be necessary to have more refined hand matrices, since uncertainty between a pair of threes and a pair of fours can suddenly become fatal. Thus, table stakes poker provides a much more severe test environment than fixed-size betting.

## 6.3 A DYNAMIC BAYESIAN NETWORK

In addition to the improvements described in § 6.1, we anticipate using a dynamic Bayesian network (DBN) to model more effectively the interrelation between rounds of play. Currently, we use a succession of four distinct Bayesian networks, one per round. In a DBN model for poker, each round would correspond to a single time which would contain the nodes *BPP Current*, *OPP Current*, *OPPUpcards* and *OPP Action*. The probability of winning would then depend on the current hand types for the final time slice (which correspond to the final hand types in the current BN).

## 7 CONCLUSION

Poker is the quintessential game combining physical probabilities (the randomness introduced by shuffling) with epistemic probabilities (the unknown values of hidden cards) with the uncertainties of assessing the opponent's psychology (propensity to bluff, strategic intentions). It appears to be an ideal domain for investigating the application of Bayesian networks. The game is too complex to model precisely with Bayesian networks, at least with the technology available to us; however, we have developed a simple and practical Bayesian network and have demonstrated its effectiveness against two reasonable computational alternatives as well as against non-expert amateur play. We expect to be able to improve the standard of play of the Bayesian player substantially towards the level achieved by human experts using the techniques outlined above.

## Invitation

Anyone wishing to test our poker program may telnet to `indy13.cs.monash.edu.au` and login as "poker" with the password "maverick".

## Acknowledgements

We acknowledge the assistance of Jon Oliver, Tam Lien, Aidan Doyle, Jamie Scuglia and Scott Thompson.